\documentclass{article}
\usepackage{ijcai17}
\usepackage{hyperref}

\usepackage{times}

\usepackage{amsmath,amsthm,amssymb, mathtools}
\usepackage{bm}
\newcommand{\bx}{\bm{x}}

\newcommand{\bX}{\mathbf{X}}

\newcommand{\tTheta}{\tilde{\Theta}}
\newcommand{\tX}{\tilde{\mathbf{X}}}

\newcommand{\ty}{\tilde{y}}

\DeclarePairedDelimiter\abs{\lvert}{\rvert}%
\DeclarePairedDelimiter\norm{\lVert}{\rVert}%
\makeatletter
\let\oldabs\abs
\def\abs{\@ifstar{\oldabs}{\oldabs*}}
\let\oldnorm\norm
\def\norm{\@ifstar{\oldnorm}{\oldnorm*}}
\makeatother

\DeclareMathOperator*{\argmax}{argmax} 

\usepackage{bbm}
\usepackage{xcolor}

\newcommand\bovermat[2]{%
  \makebox[0pt][l]{$\smash{\overbrace{\phantom{%
    \begin{matrix}#2\end{matrix}}}^{\text{#1}}}$}#2}

\usepackage{enumitem}
\newenvironment{prettyitem}[1]{
\vspace{-0.5mm}
\begin{itemize}[leftmargin=#1]
\setlength \itemsep{-0.3em}}
{\end{itemize}}

\usepackage{subcaption}
\usepackage{graphicx}
\usepackage{capt-of}
\usepackage{hyperref}
\usepackage{natbib}

\usepackage{multirow}

\newcommand{\specialcell}[2][c]{%
  \begin{tabular}[#1]{@{}c@{}}#2\end{tabular}}

\usepackage{algorithm}
\usepackage{algorithmicx}
\usepackage{algpseudocode}

\title{ An Efficient Pseudo-likelihood Method for \\Sparse Binary Pairwise Markov Network Estimation}
\author{Sinong Geng, Zhaobin Kuang, and David Page\\ 
University of Wisconsin\\
sgeng2@wisc.edu, zkuang@wisc.edu, page@biostat.wisc.edu}

\begin{document}

\maketitle

\begin{abstract}
The pseudo-likelihood method~\citep{hofling2009estimation} is one of the most popular algorithms for learning sparse binary pairwise Markov networks. In this paper, we formulate the $L_1$ regularized pseudo-likelihood problem as a sparse multiple logistic regression problem. In this way, many insights and optimization procedures for sparse logistic regression can be applied to the learning of discrete Markov networks. Specifically, we use the coordinate descent algorithm for generalized linear models with convex penalties \citep{friedman2010regularization}, combined with strong screening rules \citep{tibshirani2012strong}, to solve the pseudo-likelihood problem with $L_1$ regularization. Therefore a substantial speedup without losing any accuracy can be achieved. Furthermore, this method is more stable than the node-wise logistic regression approach on unbalanced high-dimensional data when penalized by small regularization parameters. Thorough numerical experiments on simulated data and real world data demonstrate the advantages of the 	proposed method. 
\end{abstract}

\section{Introduction}
Markov networks are a class of probabilistic graphical models that is widely applicable to many areas like image processing \citep{mignotte2000sonar}, multiple testing \citep{liu2016multiple} and computational biology \citep{friedman2004inferring}. In a Markov network, the conditional independence relationships among random variables are illuminated by the structure of the network. The most difficult challenge in estimating binary pairwise Markov networks (BPMNs) has always been dealing with the intractable computation related to log-likelihoods, which makes the learning process an NP-hard problem. Therefore, in the literature, various methods have been proposed to \emph{approximate} the log-likelihood function instead of using exact estimation. \citet{hinton2002training} built a contrastive divergence algorithm by directly estimating the derivative of log-likelihood for a discrete Markov network. \citet{wainwright2007high} considered the neighborhood recovery for each variable separately and proposed the node-wise logistic regression (NLR) method. In \citet{hofling2009estimation}, pseudo-likelihood (PL) was proposed as an approximation to the log-likelihood of a penalized BPMN. 

According to \citet{hofling2009estimation}, the PL method is one of the most competitive methods, with faster speed and higher accuracy than the other approaches. However, because of the development in the optimization for other competing methods, the PL method solved by an existing implementation, the \texttt{BMN} package \citep{BMNpdf21:online}, became the slowest among many learning methods for discrete Markov networks \citep{viallon2014empirical}; there still remains demand for a more efficient optimization procedure for PL.

Meanwhile, $L_1$-regularized logistic regression (LR), as one of the most widely used generalized linear models (GLMs), has sophisticated implementations that can deliver solutions efficiently. The state-of-the-art implementation of $L_1$-regularized GLMs leverages coordinate descent \citep{friedman2010regularization} with variable screening \citep{tibshirani2012strong}, and has become a building block for many other sparse learning problems like the ones in \citet{zhao2014pathwise}, \citet{kuangbaseline} and so on.

As noticed by some researchers, there have been some potential similarities between the objective functions of PL and LR. \citet{hofling2009estimation} pointed out that the PL model is related to LR problems. \citet{yang2011use} considered PL as an LR problem with symmetric constraints. \cite{guo2010joint} pointed out the equivalence of the objective functions of the two problems and provided an optimization algorithm based on this relationship. However the advantages of treating the sparse PL model as a regularized LR problem has been neither fully exploited nor emphasized enough. As a result, in a recent empirical study that compares multiple estimation mehtods for BPMNs \citep{viallon2014empirical}, optimization procedures with suboptimal efficiency are still considered and benchmarked to solve PL problems.

By posing PL as an LR problem in the context of learning an $L_1$-regularized BPMN, a much faster alternative to the optimization of PL can be attained. Specifically our work in this paper is summarized as follows:

\begin{itemize}[leftmargin=*]
\item With the relationship between PL and LR, we consider an optimization procedure using the coordinate descent algorithm and variable screening to solve the $L_1$ regularized PL problem.  The procedure in question can be conveniently implemented via the state-of-the-art optimization algorithm for learning sparse generalized linear models: \texttt{glmnet}. Thus, the procedure is called PLG (pseudo-likelihood using glmnet). Achieving a dramatic speedup without losing \emph{any} accuracy, PLG substantially outperforms the highly visible implementation of PL \citep{BMNpdf21:online}. 
\item Unlike the NLR approach, the PLG procedure maintains the efficiency even when dealing with unbalanced high-dimensional data penalized by small regularization parameters. We provide insights and numerical experiments to demonstrate the superior stability of the PLG method.     
\end{itemize}	   

\section{Background}
\label{sec:background}
To motivate the optimization approach for PL, we first review the background knowledge about BPMNs. 

We consider a $p$-dimensional binary observation $\bx = (x_1, x_2, \ldots, x_p)^\top\in \{0,1\}^p$. In a BPMN, the distribution of $\bx$ is associated with a network with vertex set $V = \{1, 2, \ldots, p\}$ and edge set $E \subseteq V \times V$. Accordingly, based on the ground truth parameter $\Theta^*$, a $p \times p$ symmetric matrix given as: 
\begin{equation*}
\Theta^* = 
\begin{bmatrix}
    \theta^*_{11} & \theta^*_{12} &\dots  & \theta^*_{1p} \\
    \theta^*_{21} & \theta^*_{22} & \dots  & \theta^*_{2p} \\
    \vdots & \vdots  & \ddots & \vdots \\
    \theta^*_{p1} & \theta^*_{p2} & \dots  & \theta^*_{pp}
\end{bmatrix}
 = \left[ \theta^*_{ij}\right]_{p \times p},
\end{equation*}
the joint probability mass function  is defined as:
\begin{equation}
\begin{split}
\label{sq:pdf}
\text{P}_{\Theta^*}(\bx) = \exp\Bigg(\sum_{s\in V} \theta^{*}_{ss} x_s + \sum_{(s,t) \in E}\theta^*_{st} x_s x_t - \Psi(\Theta^*)\Bigg),
\end{split}
\end{equation}
where $\Psi(\Theta)$, for any symmetric $\Theta = \left[ \theta_{ij}\right]_{p \times p} $, denotes the log-partition function defined as:
\begin{equation*}
\begin{split}
\Psi(\Theta) = \log\Bigg[\sum_{\bx \in\{0,1\}^p }\exp\Bigg(\sum_{t\geqslant s \geqslant 1}\theta_{st} x_s x_t \Bigg) \Bigg].
\end{split}
\end{equation*}
 
In order to estimate (\ref{sq:pdf}), people consider the $L_1$-regularized log-likelihood function for the BPMN:
\begin{equation}
\label{sq:loglikelihood}
\mathcal{L}(\Theta ;\bX) = \sum^{p}_{t\geqslant s \geqslant 1} \theta_{st} (\bX\bX^\top)_{st} - N\Psi(\Theta) - \frac{N \lambda}{2} \sum_{s \neq t}\abs{\theta_{st}},
\end{equation} 
where $(\bX\bX^\top)_{st}$ denotes the element in the $s^{th}$ row and the $t^{th}$ column of $\bX\bX$, given $N$ independent and identically distributed samples $\bX = (\bx_1, \bx_2, \ldots, \bx_N)^\top = \left[ \bX_{ij}\right]_{N \times p}$. The goal of dealing with (\ref{sq:loglikelihood}) is to estimate $\Theta^*$ with the optimal $\Theta$ given as: 
\begin{equation*}
\Theta_{\bX}^* = \argmax_{\Theta}\mathcal{L}(\Theta,\bX),
\end{equation*}
which can be can be extremely challenging because of the intractable log-partition function, $\Psi(\Theta)$. Therefore, instead of maximizing (\ref{sq:loglikelihood}), the PL method considers the $L_1$-regularized pseudo-likelihood function \citep{hofling2009estimation},
\begin{equation}
\begin{multlined}
\label{sq:sample pseudo-likelihood}
\hat{\mathcal{L}}(\Theta;\bX) = 
\sum_{n=1}^N 
\sum^p_{s = 1}\Bigg[\bX_{ns}\Bigg( \theta_{ss} + \sum_{t\neq s} \bX_{nt} \theta_{st} \Bigg) \\
- \Psi_s(\bx_n, \Theta) \Bigg]
- N \lambda \sum_{t > s}\abs{\theta_{st}}, 
\end{multlined}
\end{equation}
as a replacement for the penalized log-likelihood function (\ref{sq:loglikelihood}) and solves for 
\begin{equation}
\label{sq:PL}
\hat{\Theta}_{\bX}^* = \argmax_{\Theta}\hat{\mathcal{L}}(\Theta,\bX)
\end{equation}
to estimate $\Theta^*$.
Here, $\Psi(\Theta)$ is replaced by the much simpler 
\begin{equation}
\label{sq: fans}
\Psi_s(\bx; \Theta) = \log \Bigg[1 + \exp \Bigg(\theta_{ss} + \sum_{t \neq s} x_t \theta_{st}\Bigg)\Bigg].
\end{equation}  

Compared with exact methods that solve (\ref{sq:loglikelihood}), the PL method that solves (\ref{sq:sample pseudo-likelihood}) is shown to be more efficient without sacrificing too much accuracy \citep{hofling2009estimation}.

There is also a connection between (\ref{sq:sample pseudo-likelihood}) and the objective function of the NLR algorithm, which separately maximizes 
\begin{equation}
\label{sq:node-wise}
\begin{multlined}
\sum_{n=1}^N
\Bigg[\bX_{ns}\Bigg( \theta_{ss} + \sum_{t\neq s} \bX_{nt} \theta_{st} \Bigg)-  \Psi_s(\bx_n, \Theta) \Bigg]
\\- N \lambda \sum_{t > s}\abs{\theta_{st}}
\end{multlined},
\end{equation}
for all $s \in V$. In fact, (\ref{sq:sample pseudo-likelihood}) is just the sum of (\ref{sq:node-wise}) on $s$'s. Specifically, \cite{ravikumar2010high} considered maximizing (\ref{sq:node-wise}) as an $L_1$-regularized LR problem with response 
\begin{equation}
\label{sq:ny}
y_s = (\bX_{1,s}\, \bX_{2,s}\, \ldots\, \bX_{N,s})^\top. 
\end{equation}

\section{Conversion from a Pseudo-likelihood Problem to a Sparse Logistic Regression Problem}
\label{sec: relationship}
We now demonstrate the relationship between the objective functions in PL and LR by transforming (\ref{sq:sample pseudo-likelihood}) into a logistic loss function with parameter $\tTheta$, design matrix $\tX$, and response $\ty$, which are defined subsequently.

We first define parameter $\tTheta$. Since in LR problems the parameter is a vector instead of a matrix like $\Theta$, we redefine the parameter in BPMNs by stacking the upper triangular elements of $\Theta$ column by column to a vector and appending the diagonal elements to the end. Thus we have

\begin{equation}
\label{sq:vector theta}
\begin{matrix}
 \tTheta
 =
 \begin{pmatrix}
 \bovermat{upper triangular elements}{\theta_{12},  & \theta_{13}, & \theta_{23}, \ldots , \theta_{(p-1)p},} & \bovermat{diagnoal elements}{\theta_{11},& \ldots, & \theta_{pp}} \\[0.5em]

  \end{pmatrix}
  
 \end{matrix},
 \end{equation}   
where for any $(s,t) \in \{(s,t)|s \neq t,\, (s,t) \in V \times V \}$, $\theta_{st}$ is transformed into the $j_{st}^{th}$ element in $\tTheta$ with 
\begin{equation*}
j_{st} = \min(s,t)+\frac{(\max(s,t)-2)(\max(s,t)-1)}{2}.
\end{equation*} 

Then for the definition of matrix $\tX = \big[ \tX_{ij} \big]_{Np \times (m+p)}$, with $m = \frac{{p(p-1)}}{2}$ denoting the number of upper triangular elements, we review the concept of the indicator function: 
\begin{equation*} 
\mathbbm{1}{( C )} = \begin{cases}
1& C \, \text{is} \, \text{sataified}\\
0& \text{otherwise}\\
\end{cases}.
\end{equation*}
Furthermore, we define 
\begin{equation} 
\tX_{ij} = \begin{cases}
\bX_{n t}  & \exists \, (s,t) \: \text{s.t.} \: j = j_{st}\\
\mathbbm{1}{ (s=j-m) }& m+1 \leqslant j \leqslant m+p\\
0 & \text{otherwise}
\end{cases},
\label{sq:x}
\end{equation}
where $i = N(s-1)+n$, $n \in \{1,2, \ldots, N\}$. 

Finally, the response $\ty$ is defined as:
\begin{equation}
\label{sq:y}
\ty = \left(\bX_{11}, \bX_{21}, \ldots, \bX_{N1}, \ldots, \bX_{1p}, \bX_{2p},\ldots, \bX_{N p}\right)^\top.
\end{equation}

With $\tX$, $\tTheta$, and $\ty$, we can rewrite the first part (the log-likelihood) of (\ref{sq:sample pseudo-likelihood}) as:
\begin{align*}
& \ty^\top \tX \tTheta - \sum_{n = 1}^{N}\sum_{k=1}^p\Psi_k(\bx_{n}, \Theta)\\
= & \ty^\top \tX \tTheta -  \sum_{k=1}^{Np} \log\left[1 + \exp\left((\tX)_{k}^\top \tTheta\right)\right],
\end{align*}
where
\begin{equation*}
(\tX)_{k} = (\tX_{k1}, \tX_{k2}, \ldots, \tX_{k(m+p)})^\top,
\end{equation*}
and
\begin{equation*}
\sum_{n = 1}^{N}\sum_{k=1}^p\Psi_k(\bx_n, \Theta) = \sum_{k=1}^{Np} \log\left[1 + \exp\left((\tX)_{k}^\top \tTheta\right)\right]
\end{equation*}
because of (\ref{sq: fans}). Therefore, (\ref{sq:sample pseudo-likelihood}) is equal to
\begin{equation}
\label{sq:converted}
\ty^\top \tX \tTheta - \sum_{k=1}^{Np} \log\left[1 + \exp\left((\tX)_{k}^\top \tTheta\right)\right] -N \lambda \sum_{s \neq t}\abs{\theta_{st}},
\end{equation} 
which is exactly the loss function for a penalized LR problem consisting of $Np$ samples with the design matrix $\tX$, response $\ty$, and the parameter $\tTheta$. 

As a result, we have converted an $L_1$-regularized pseudo-likelihood problem into an LR problem with the objective function (\ref{sq:converted}). 

Based on the relationship established above, we can solve a spare PL problem by solving its \emph{equivalent} sparse LR problem. The consequence is that we can take advantage of the sophisticated optimization procedures for sparse LR problems to compute the solution for a PL problem efficiently. We now consider the optimization procedure for sparse PL problems based on one of the most efficient optimization algorithms for penalized LR problems, the coordinate descent algorithm \citep{friedman2010regularization}. Furthermore, we also use the initialization procedure provided by the strong screening rule \citep{tibshirani2012strong} for a further speedup. The details of PLG are presented in Algorithm~\ref{alg:plg}. 

\captionsetup[algorithm]{font=footnotesize}
\begin{algorithm}[H]
\caption{Pseudo-likelihood Method using \texttt{glmnet} (PLG)}
\begin{algorithmic}[1]
\Require $\bX$, $\lambda$.
\Ensure $\hat{\Theta}^*_{\bX}$.
\State Build $\tX$ with $\bX$ according to (\ref{sq:x}).
\State Build $\ty$ with $\bX$ according to (\ref{sq:y}).
\State Solve a sparse LR problem  for $\tTheta$ via \texttt{glmnet} with $\tilde{y}$ as the response, and $\tX$ as features under $\lambda$.
\State Transform $\tTheta$ into $\hat{\Theta}^*_{\bX}$ according to (\ref{sq:vector theta}).
\end{algorithmic}
\label{alg:plg}
\end{algorithm}
\begin{table*}
\centering
\begin{tabular*}{0.9\textwidth}{@{\extracolsep{\fill} }c c c c c c}
\hline
& \multirow{2}{*}{Number of Vertices ($p$)} &\multicolumn{4}{ c }{Edge Generation Probability ($\mathbf{P}$)} \\ \cline{3-6} 
 && 0.2 & 0.3 & 0.4 & 0.5\\ 
\hline
\multicolumn{1}{ c  }{\multirow{5}{*}{BMN \& PLG}} & 5 & 0.01566 &0.0117& 0.0189&0.02016 \\
\multicolumn{1}{ c  }{}
 & 10 & 0.01971 &0.0144&0.01863&0.01728\\
 \multicolumn{1}{ c  }{}
 & 15 & 0.018 &0.01408&0.0171&0.012\\
 \multicolumn{1}{ c  }{}
 & 20 & 0.0135 &0.006975&0.0135&0.009\\ 
 \multicolumn{1}{ c  }{}
 & 25 & 0.0108&0.00702&0.0108&0.0072\\ \hline
\multicolumn{1}{ c  }{\multirow{5}{*}{NLR}} & 5 & 0.00783 &0.00585& 0.00945&0.01008\\
\multicolumn{1}{ c  }{}
 & 10 & 0.009855 &0.0072&0.009315&0.00864\\
\multicolumn{1}{ c  }{}
 & 15 & 0.009 &0.0054&0.00855&0.006\\
\multicolumn{1}{ c  }{}
 & 20 & 0.00675 &0.003488&0.00675&0.0045\\ 
\multicolumn{1}{ c  }{}
 & 25 & 0.0054&0.00351&0.0054&0.0036\\ \hline 
\end{tabular*}
\caption {$\lambda$'s selected by StARS for different methods in different networks} 
  \label{tab:lambda}
\end{table*}
\begin{figure*}
\centering
\begin{subfigure}{0.24\textwidth}
\centering
\includegraphics[scale=0.2]{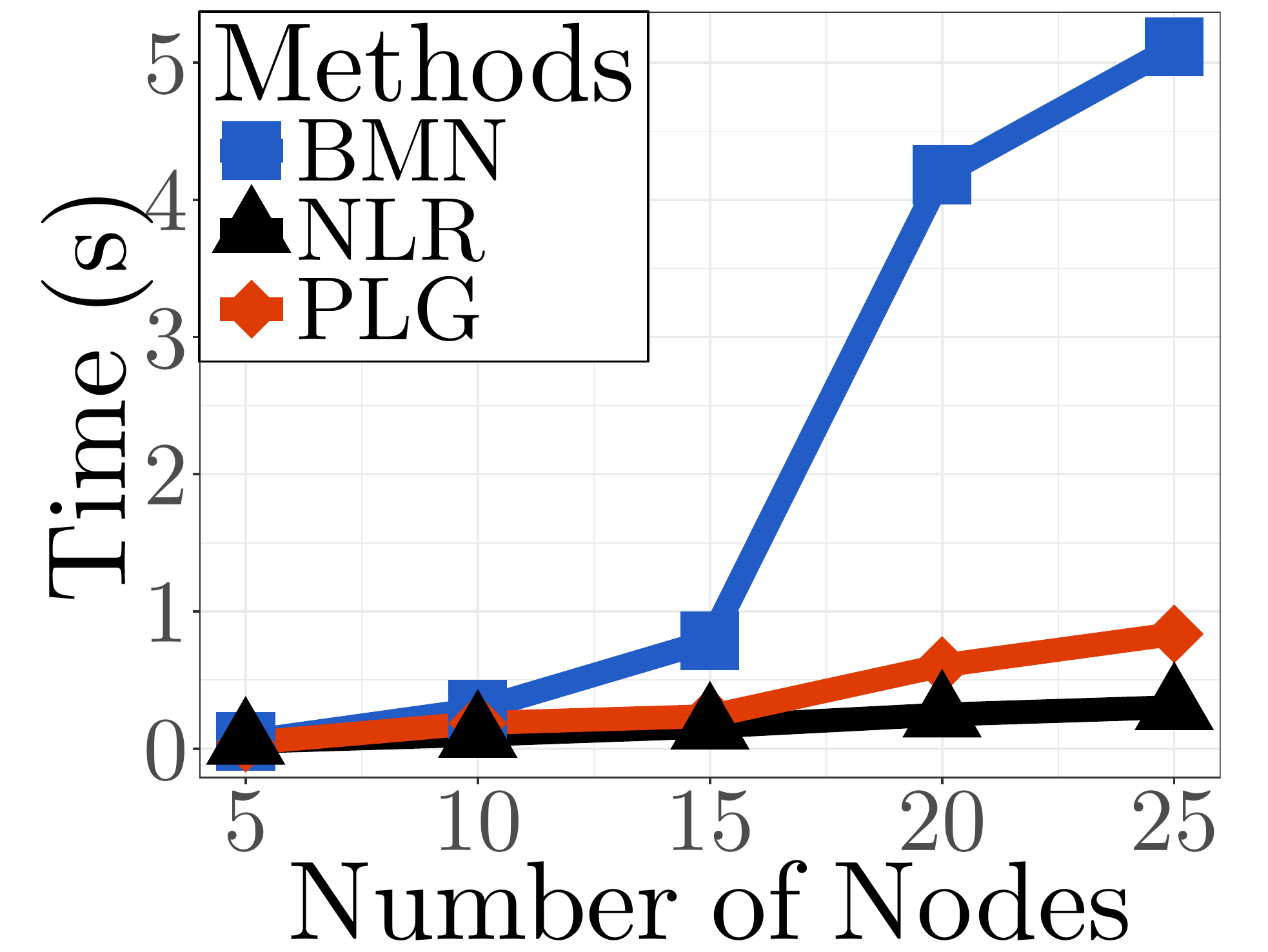}
\caption{$\mathbf{P} = 0.2$}
\end{subfigure}
\begin{subfigure}{0.24\textwidth}
\centering
\includegraphics[scale=0.2]{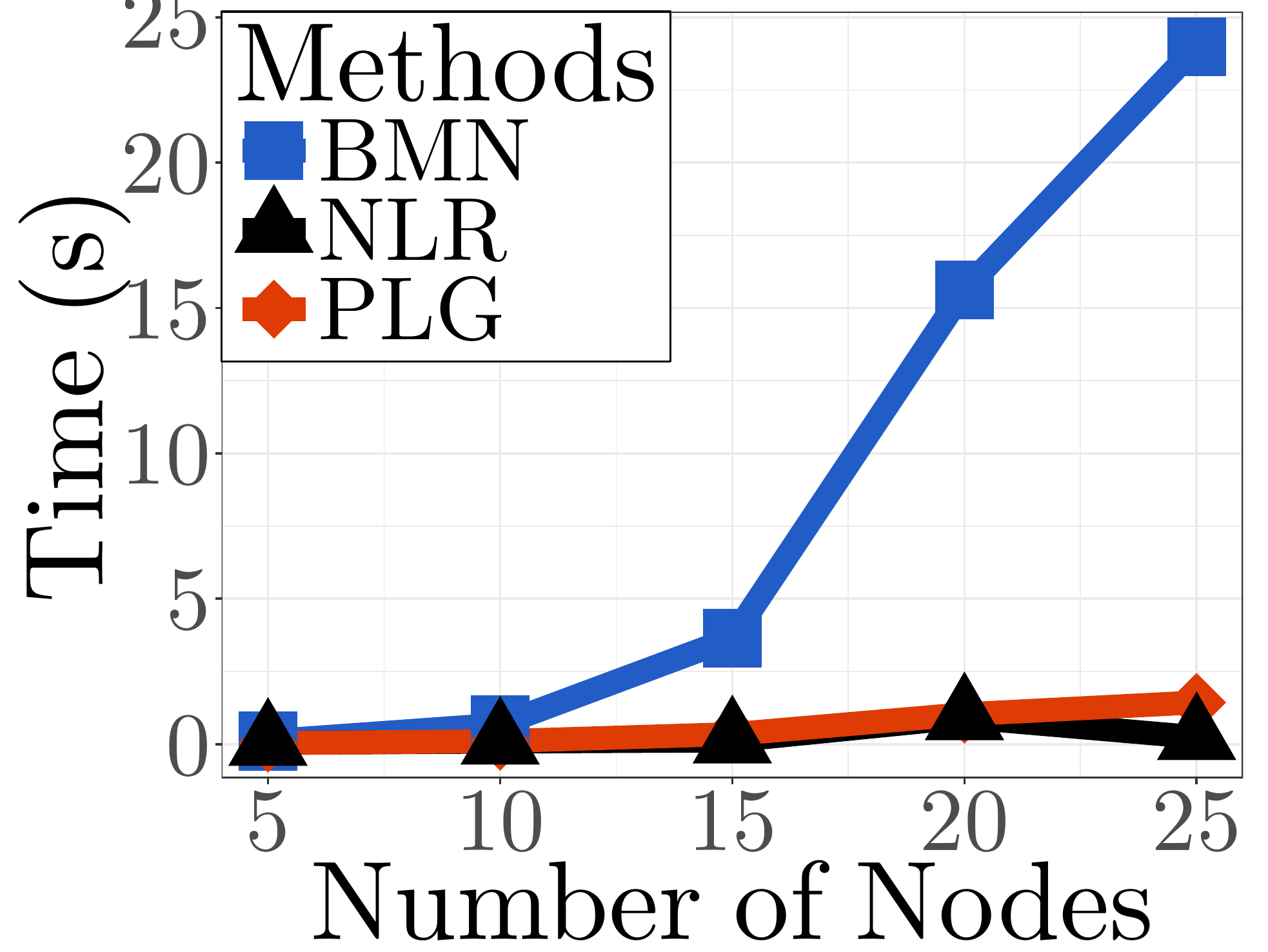}
\caption{$\mathbf{P} = 0.3$}
\end{subfigure}
\begin{subfigure}{0.24\textwidth}
\centering
\includegraphics[scale=0.2]{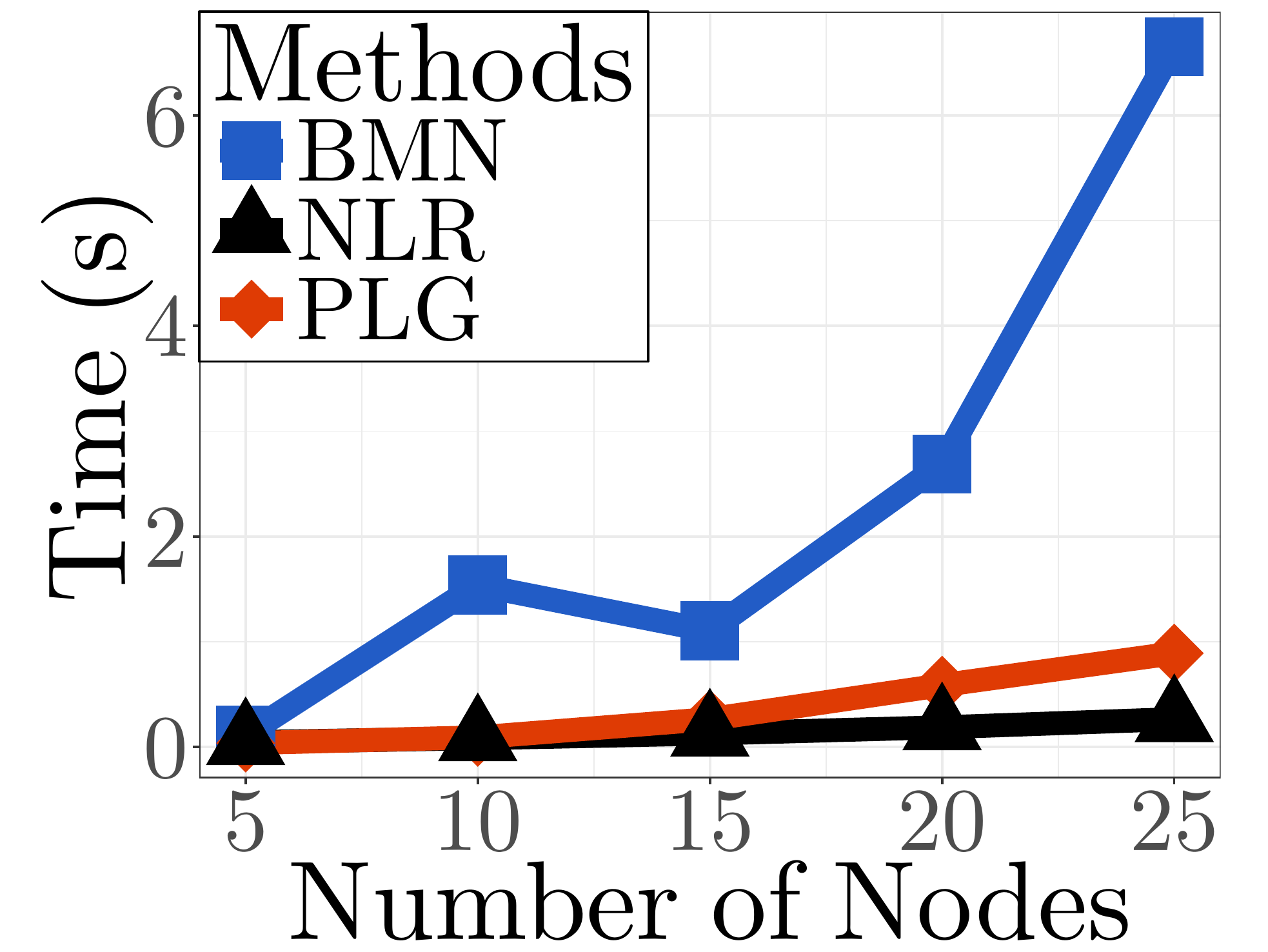}
\caption{$\mathbf{P} = 0.4$}
\end{subfigure}
\begin{subfigure}{0.24\textwidth}
\centering
\includegraphics[scale=0.2]{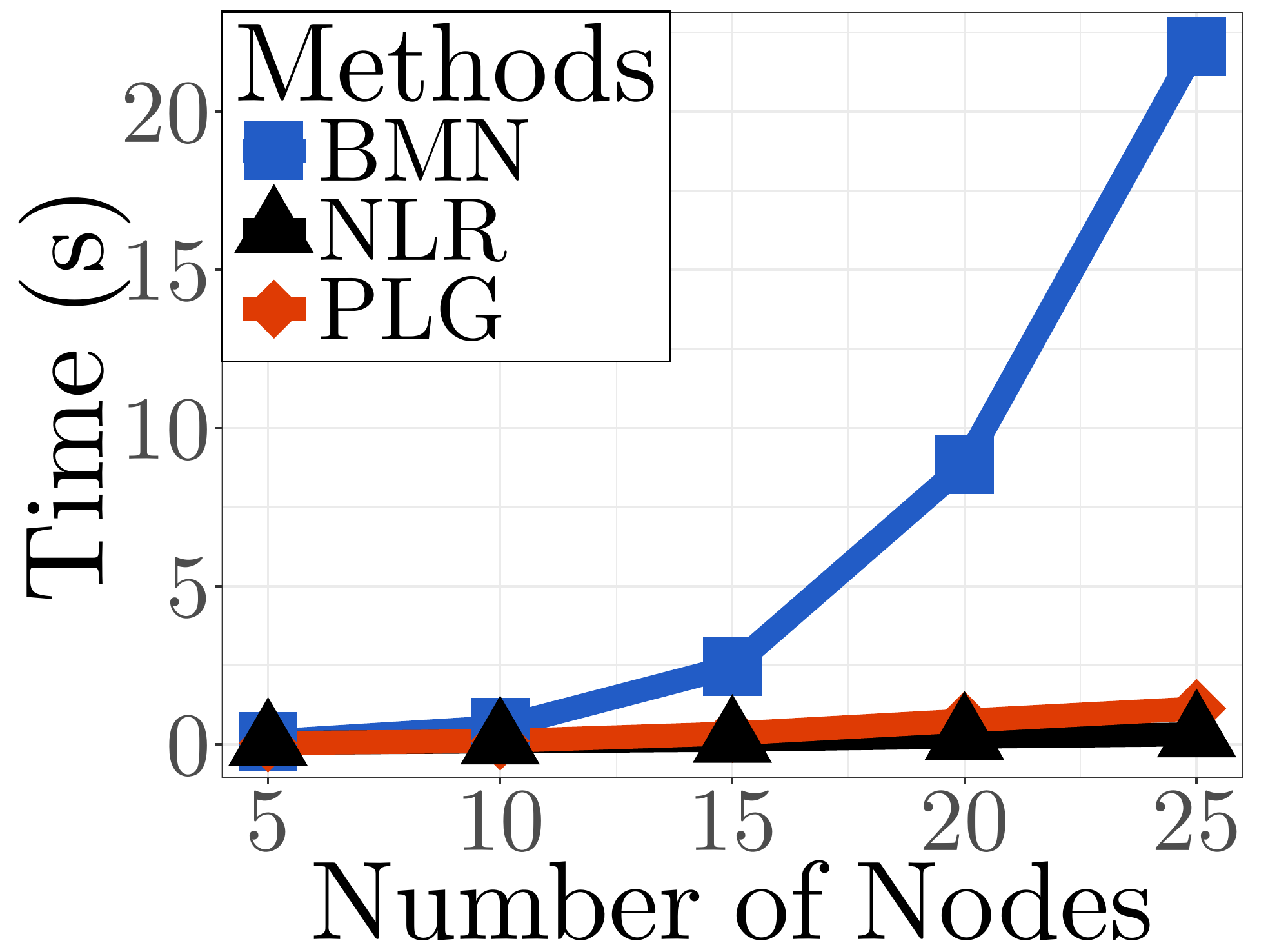}
\caption{$\mathbf{P} = 0.5$}
\end{subfigure}
\caption{Computation time of BMN, NLR and PLG evaluated on graphs with edge generation probability of $0.2$, $0.3$, $0.4$ and $0.5$.}
\label{fig:time}
\end{figure*}

\section{PLG versus NLR When Learning from Unbalanced Data}
\label{sec:analysis on stability}
In this section, we compare the performance of the PLG method with the NLR algorithm when dealing with unbalanced high-dimensional data under small regularization parameters.

When the data are unbalanced, some columns of $\bX$ will be predominantly $0$'s or $1$'s. Since NLR uses each column of $\bX$ in turn as a response, columns with predominantly $0$'s or $1$'s will also serve as responses in some spare LR models. Unfortunately, in this situation, we observe that NLR will be extremely slow and even may fail to converge. To make things worse, since the optimal solution gets denser with the decrease of the regularization parameter, the efficiency of NLR further deteriorates when penalized by small $\lambda$'s.

On the contrary, PLG is more stable in this situation. In the PLG method, the response is chosen to be a longer vector $\ty$ in (\ref{sq:y}), consisting of all the elements in $\bX$ instead of just one column. Therefore, the unbalanced nature of one column in $\bX$ cannot have a huge effect on $\ty$. That is to say, while NLR is dealing with highly unbalanced responses, PLG is still solving an ordinary LR problem with a relatively balanced response, and thus maintains high efficiency even in the context of unbalanced high-dimensional data penalized by small regularization parameters. We will illustrate this phenomenon in detail in Section \ref{sec:stability} by experiments.

\section{Numerical Experiments}
In this section, we compare the empirical performance of the PLG algorithm with those of the original implementation of the PL method by \texttt{BMN} (BMN) \citep{BMNpdf21:online} and the NLR algorithm \citep{viallon2014empirical}. First, we show that, as a more efficient optimization method, PLG achieves the same objective function value as BMN with a much faster speed. We also apply NLR and PLG to unbalanced high-dimensional data for a deeper understanding of the relative efficiency of the two methods. Furthermore, we evaluate the structure learning performances  of the three methods using  receiver operating characteristic (ROC) curves as our evaluation metric. ROC curves of the three methods are shown to be very similar on simulated datasets. Finally, performances of the three methods on real world data are also investigated using the senator voting record \citep{USSenate38:online} dataset.

Except BMN, the existing best implementation of PL methods, we are only comparing PLG with NLR, while many estimation methods for sparse BPMNs have been proposed in the literature \citep{banerjee2008model, yang2011use, anandkumar2012high}. We believe that the comparison between the performance of PLG and that of NLR is representative because NLR has been empirically shown to have the highest efficiency compared with other competing methods especially when dealing with high-dimensional data \citep{viallon2014empirical}. Furthermore, as shown in Section \ref{sec: relationship}, NLR also uses a kind of pseudo-likelihood to approximate log-likelihood and thus has a similar objective function to that of PLG, making the contrast between PLG and NLR a natural comparison.   

\subsection{Simulated Data Generation}  
We use a procedure similar to that in  \cite{hofling2009estimation} to generate the ground truth parameter $\Theta^*$ and the synthetic datasets.
\begin{prettyitem}{*}
\item The number of the vertices, $p$, in the ground truth network, is chosen to be $5$, $10$, $15$, $20$ or $25$. Each element of $\Theta$ is drawn randomly to be non-zero, with edge generation probability $\mathbf{P}\in \{0.2, 0.3, 0.4, 0.5\}$. And the non-zero elements have a uniform distribution on $[-1,1]$.
\item 1000 samples are generated by Gibbs sampling with 1000 burn-in steps.
\item The results reported from Section \ref{sec:optimization} to Section \ref{sec:stability} are averages of 20 trials.
\end{prettyitem}

\subsection{Model Selection}
\label{sec:lambda}
Before we proceed to compare the efficiency and accuracy of BMN, NLR, and PLG, we conduct model selection to find the best regularization parameters (best representatives) for the three methods in different networks. We use StARS, a stability-based regularization parameter selection method to high dimensional inference for undirected graphs \citep{liu2010stability}, to determine the $\lambda$ that achieves the best balance between the edge selection stability and the network sparsity. In addition, since BMN and PLG have the same objective function, it's reasonable to use the same $\lambda$'s for them and the $\lambda$'s for NLR should be half of those in BMN \citep{hofling2009estimation}. In detail, for different networks and methods, the $\lambda$'s selected are summaried in Table \ref{tab:lambda}. 
\begin{table*}
\centering
\begin{tabular*}{0.9\textwidth}{@{\extracolsep{\fill} } c c c c c c}
\hline
& \multirow{2}{*}{Number of Vertices ($p$)} &\multicolumn{4}{ c }{Edge Generation Probability ($\mathbf{P}$)} \\ \cline{3-6} 
 && 0.2 & 0.3 & 0.4 & 0.5\\ 
\hline
\multicolumn{1}{ c  }{\multirow{5}{*}{\specialcell{Relative \\Difference ($\epsilon$)}}} & 5 & $8.43 \times 10^{-4}$ &$9.932 \times 10^{-4}$& $8.241 \times 10^{-4}$&$12.601 \times 10^{-4}$ \\
\multicolumn{1}{ c  }{}
 & 10 & $12.601 \times 10^{-4}$ &$16.208 \times 10^{-4}$&$19.771 \times 10^{-4}$&$16.924 \times 10^{-4}$\\
 \multicolumn{1}{ c  }{}
 & 15 & $19.632 \times 10^{-4}$ &$23.136 \times 10^{-4}$&$26.934 \times 10^{-4}$&$23.749 \times 10^{-4}$\\
 \multicolumn{1}{ c  }{}
 & 20 & $27.348 \times 10^{-4}$ &$34.438 \times 10^{-4}$&$32.021 \times 10^{-4}$&$39.807 \times 10^{-4}$\\ 
 \multicolumn{1}{ c  }{}
 & 25 & $35.123 \times 10^{-4}$&$39.441 \times 10^{-4}$&$40.867 \times 10^{-4}$&$52.165 \times 10^{-4}$\\ \hline
\end{tabular*}
\caption {Relative differences (\ref{sq:reDifference}) between the parameter achieved by BMN and PLG} 
  \label{tab:reDifference}
\end{table*}
\begin{figure*}
\centering
\begin{subfigure}{0.24\textwidth}
\centering
\includegraphics[scale=0.15]{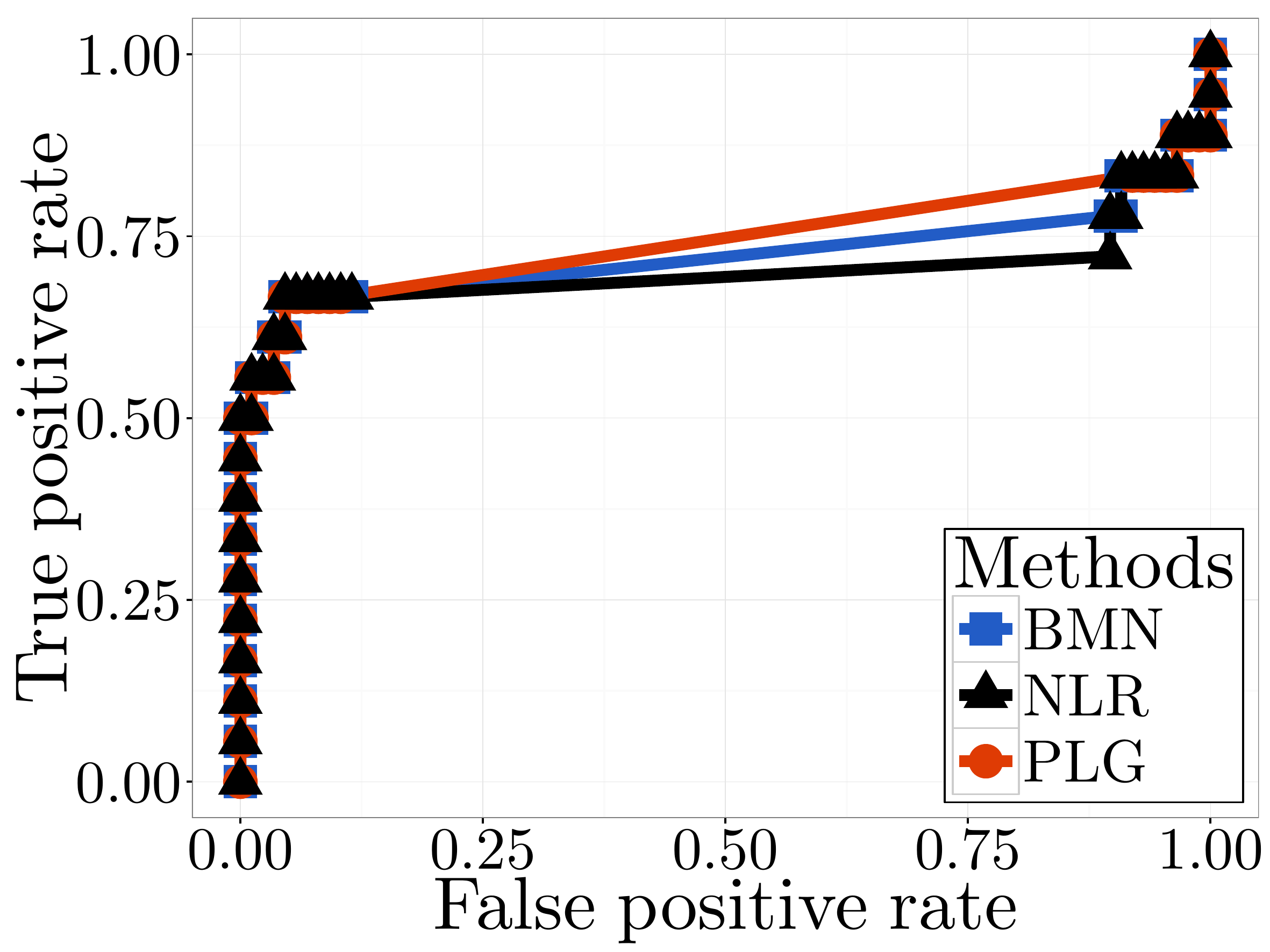}
\caption{$\mathbf{P} = 0.2$}
\end{subfigure}
\begin{subfigure}{0.24\textwidth}
\centering
\includegraphics[scale=0.15]{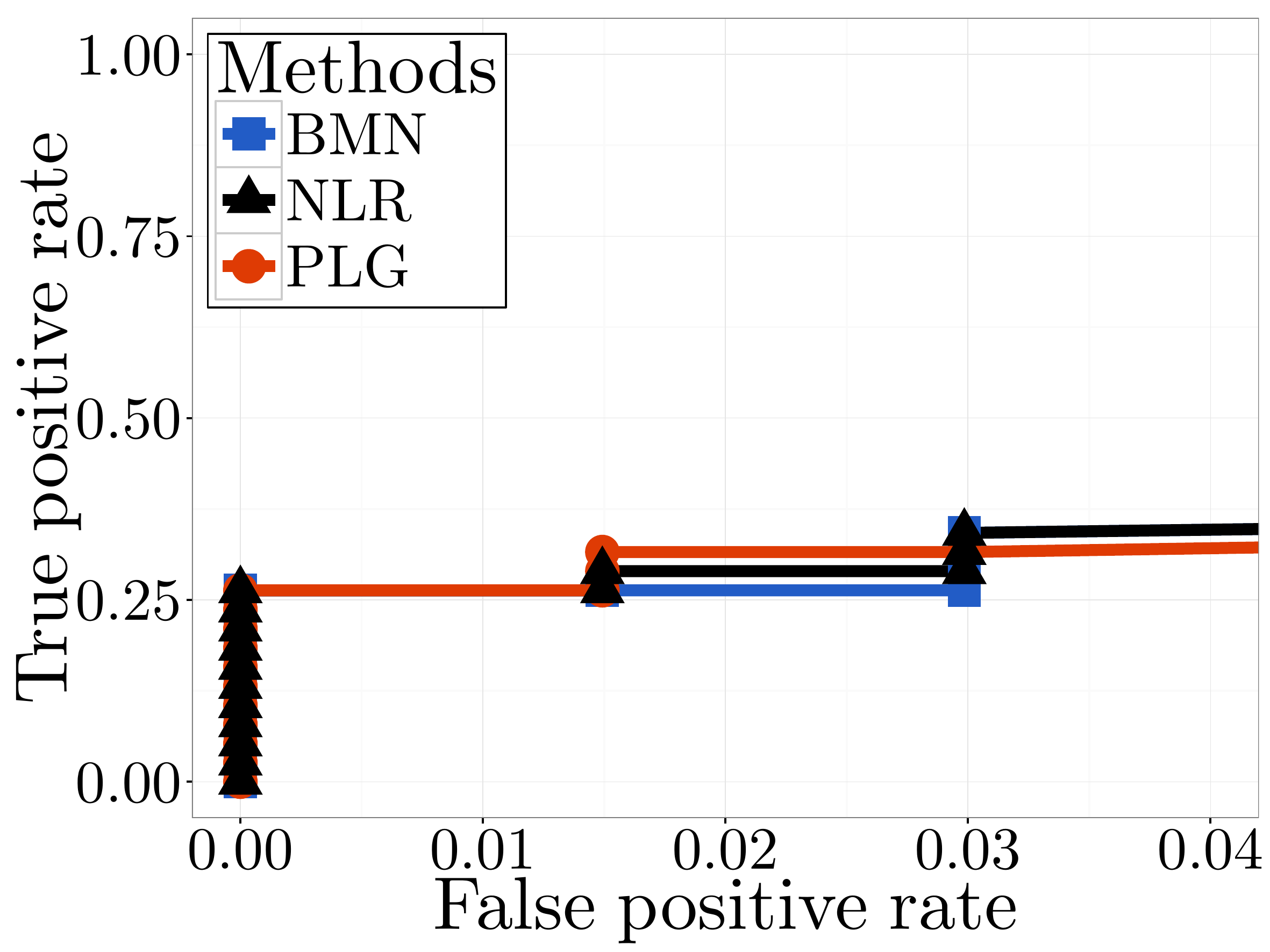}
\caption{$\mathbf{P} = 0.3$}
\end{subfigure}
\begin{subfigure}{0.24\textwidth}
\centering
\includegraphics[scale=0.15]{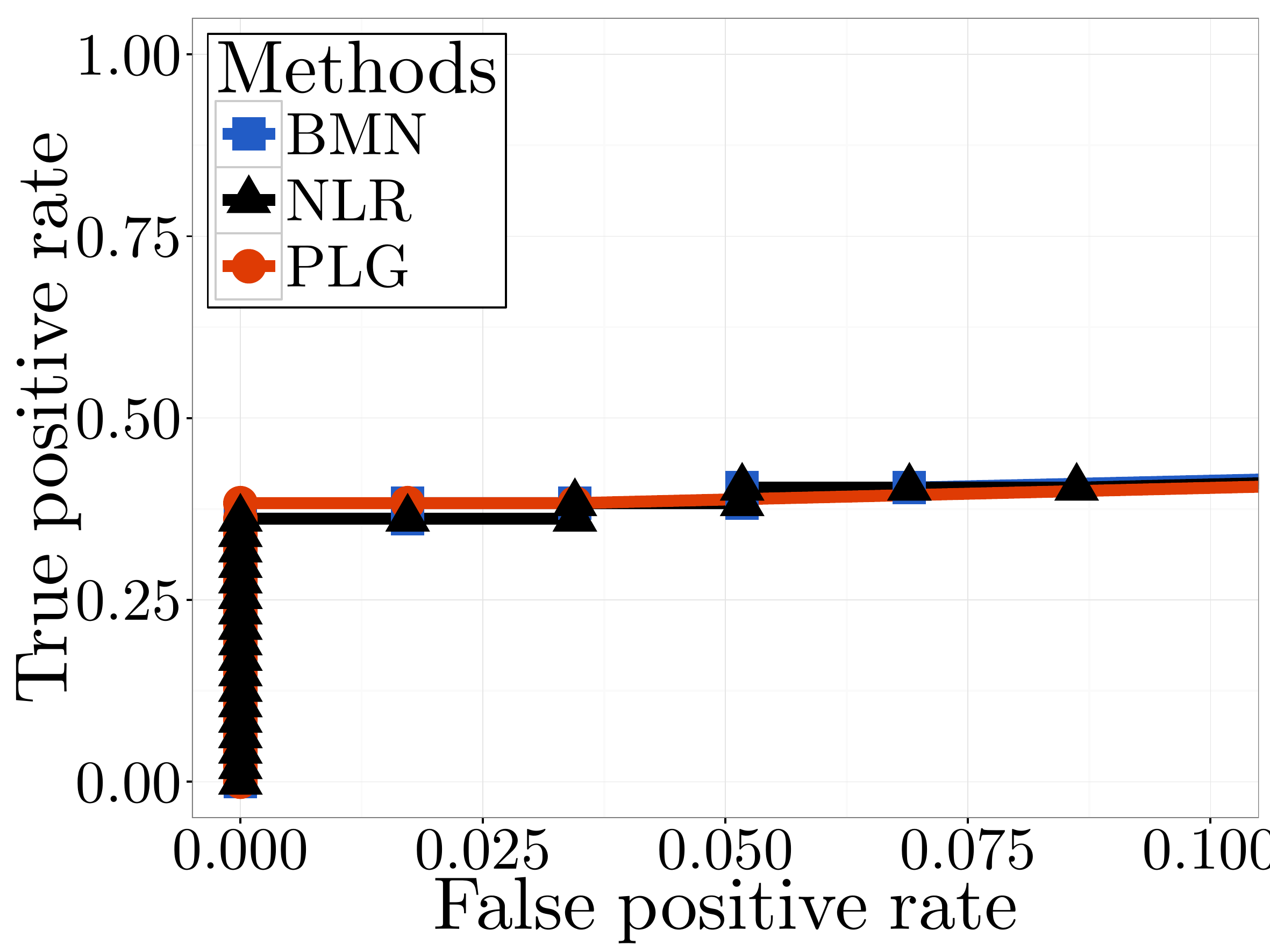}
\caption{$\mathbf{P} = 0.4$}
\end{subfigure}
\begin{subfigure}{0.24\textwidth}
\centering
\includegraphics[scale=0.15]{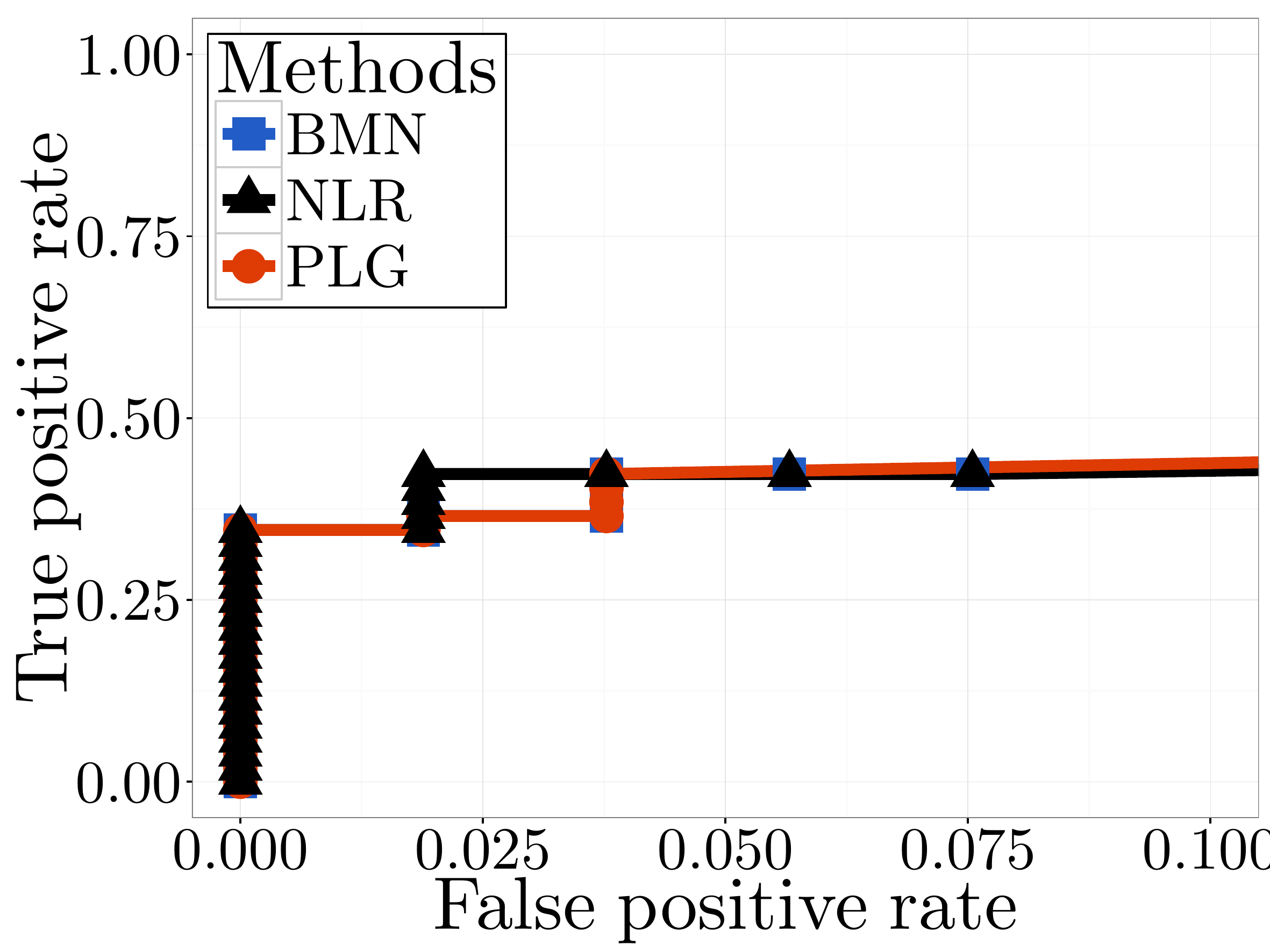}
\caption{$\mathbf{P} = 0.5$}
\end{subfigure}
\caption{False positive rate versus true positive rate for structure estimation, evaluated on graphs with edge generation probability of $0.2$, $0.3$, $0.4$ and $0.5$.}
\label{fig:ROC}
\end{figure*}
\begin{figure}[t]
\centering
\includegraphics[scale=0.35]{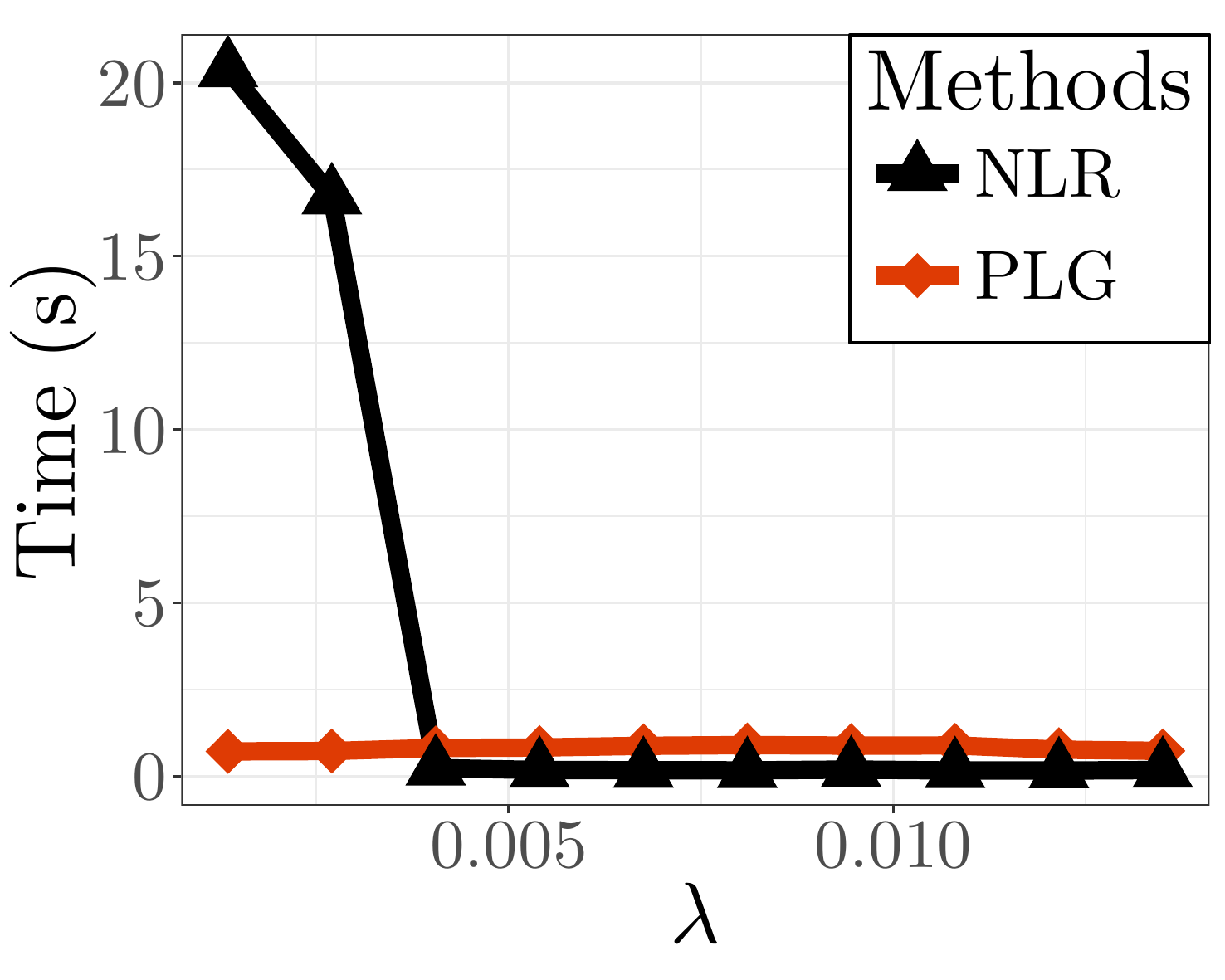}
\caption{Computation time under different $\lambda$ with $\mathbf{P} = 0.3$}
\label{fig:stability}
\end{figure}

\subsection{Efficiency}  
\label{sec:optimization}
We now compare the efficiency among BMN, NLR and PLG. Using the $\lambda$'s chosen by the model selection procedure in Section \ref{sec:lambda}, we apply the three methods to datasets generated by BPMNs with different numbers of vertices and edge generation probability $\mathbf{P}$s. The computation time of the learning process \emph{after} the selection of $\lambda$'s is reported in Figure \ref{fig:time}. 

With the improvement on optimization, PLG outperforms BMN tremendously and becomes comparable to NLR in the aspect of efficiency. Furthermore, the advantages of NLR and PLG to BMN are more substantial with the scaling up of networks. This observation is consistent with the existing result that NLR performs better for high-dimensional problems \citep{viallon2014empirical}. 

In fact, the comparable performances between NLR and PLG are also reasonable, considering that they apply similar optimization approaches to similar objective functions. 

Naturally, we want to see whether PLG compromises accuracy for acceleration. To this end, we examine the difference between the parameters achieved by PLG ($\tTheta_P$) and that achieved by BMN ($\tTheta_B$). For a clear illustration of the difference, we define the relative difference $\epsilon$ between $\tTheta_P$ and $\tTheta_B$ as
\begin{equation}
\epsilon = \frac{\norm {\tTheta_P-\tTheta_B}_2}{\norm{\tTheta_B}_2}.
\label{sq:reDifference}
\end{equation}
The $\epsilon$'s for the solutions achieved by PLG and BMN in the experiments above are shown in Table (\ref{tab:reDifference}). It should be noticed that PLG achieves nearly the same parameters as BMN. In addition, the relative difference rises with the increase of $p$ since we are using the same stopping criterion in different networks. These results suggest that PLG provides a notable improvement in efficiency to PL methods without losing any accuracy.   

\begin{figure*}
\centering
\includegraphics[scale=0.8]{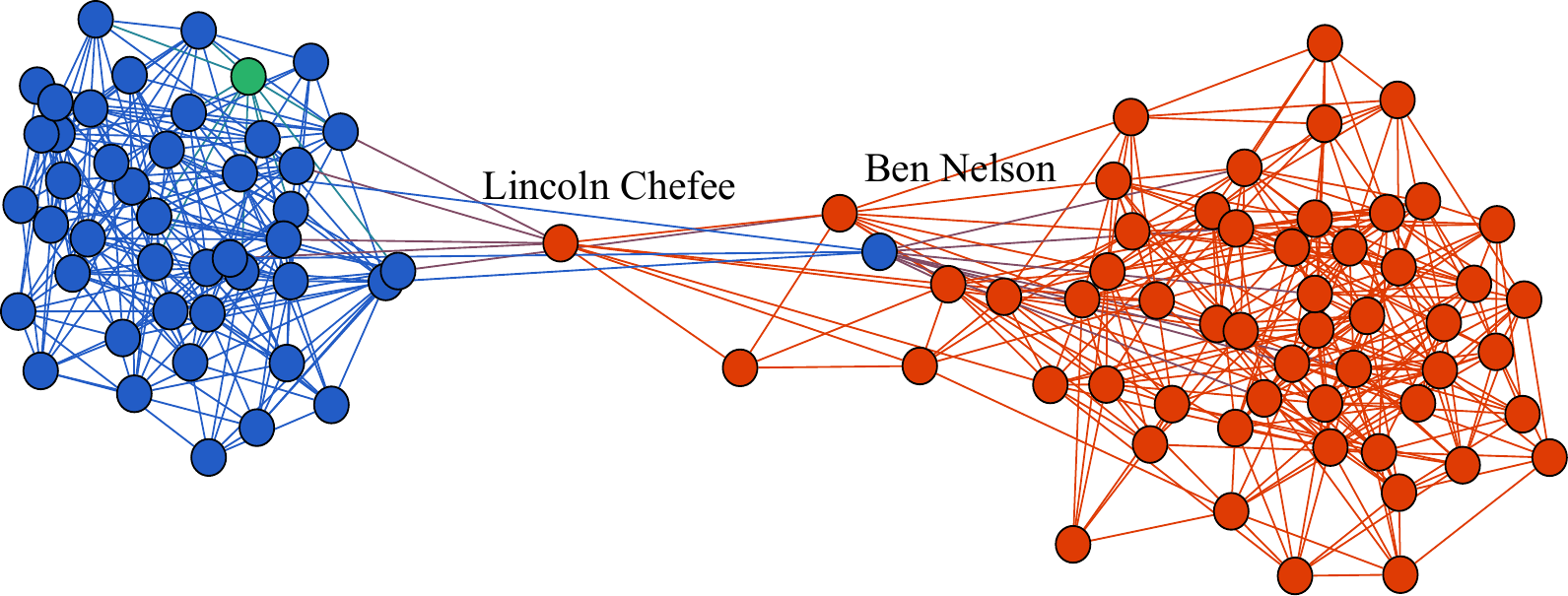}
\caption{The visualization of senator voting data: Red vertices denote Republicans, blue Democracts, and green Independent. Red arcs are among the Republicans and blue Democrats. Purple arcs represent strong ties from certain Republicans to the Democratic cluster. The figure is rendered by \texttt{Gephi} \citep{bastian2009gephi}.}\label{fig:senate} \vspace{-4mm}
\end{figure*}

\subsection{Structure Estimation}  
In order to demonstrate that PLG also has the ability to estimate the correct structure of ground truth networks we compare ROC curves of the three methods for structure estimation to networks consisting of 15 vertices with edge generation probability $\mathbf{P}\in \{ 0.2, 0.3, 0.4, 0.5\}$ respectively in Figure (\ref{fig:ROC}). Overall, we notice that all the three methods achieve nearly the same performance, and in some figures, the lines even overlap with each other, demonstrating the utility of PLG in structure estimation. This result is not surprising because of the coherence with existing empirical results in the literature \citep{hofling2009estimation, viallon2014empirical}.

\subsection{Learning from Unbalanced Data}
\label{sec:stability} 

As we mention above, with the PLG implementation, PL becomes comparable to NLR in efficiency and accuracy. In fact, PLG even outperforms NLR when dealing with unbalanced data under small $\lambda$'s. 

Analysis on this particular situation is necessary because using the joint probability mass function (\ref{sq:pdf}), it is very likely to generate extremely unbalanced data if $\theta_{ss} \neq 0$ in a BPMN. In addition, although we only use simulated data to illustrate the stability of two methods for unbalanced data with small $\lambda$'s, unbalanced data are ubiquitous in practical problems \citep{liu2014new}, especially in researches on mutations of genes. Considering the huge amount of genes and the small probability of mutations, only a few mutated samples can be observed in practice. Accordingly, samples in this kind of problems are always unbalanced and thus it's meaningful to scrutinize the extreme situation in our work.

To compare the stability of PLG and NLR with unbalanced data and small $\lambda$'s, we set the number of the vertices, $p=10$ and compare the computation time under different $\lambda$'s. In addition, we assume $\theta_{1,1}=5$, when simulating samples to generate unbalanced data. We omit the results of BMN because of the low efficiency. Results are summarized in Figure \ref{fig:stability}.

We notice that the efficiency of NLR decays immediately with small $\lambda$'s and unbalanced data while our method still maintains a fast speed, as we expect in section \ref{sec:analysis on stability}. PLG performs similarly with NLR under large $\lambda$'s, which is consistent with the results in \ref{sec:optimization}. Theses results indicate that PLG is readily available for unbalanced data.

\subsection{Real World Experiments}
In this section, performances of the three methods on real world data are investigated. 
We conduct an experiment using the senator voting record consisting of 279 samples and 100 variables in the second session of the 109th Congress\citep{USSenate38:online}. The task of interest is to investigate the clustering effect of voting consistency. That is to say, we want to find the senators who are more likely to cast similar votes on bills. Here are the details of the experiment. 
\begin{prettyitem}{*}
\item Each bill is considered as a sample and the votes from senators are features.
\item If a senator votes for one bill, the corresponding element in the sample will be denoted by 1, otherwise 0. Missing data are imputed as 0.
\item A binary pairwise Markov network is used to model the data. And we learn the network with PLG and BMN.
\item The $\lambda$ selected by the model selection procedure in Section \ref{sec:lambda} is 0.06.  
\item The vertices represent senators and the edges denote the estimated $\theta_{st}$,where $s\neq t$. Furthermore, only the edges with a positive estimated parameter are displayed.
\end{prettyitem}
The visualization of the BPMN for senator voting data is presented in Figure $\ref{fig:senate}$. First, as would be expected, senators are divided into mainly a Democratic cluster and a Republican cluster, which is roughly consistent with party memberships of the senators. Second, as a Republican, Lincoln Chafee is closer and has more connection to Democrats. In fact, he joined the Democrats in 2008 \citep{wiki:Lincoln_Chafee} and our model can detect his democratic-leaning voting pattern. Third, since Ben Nelson is ``one of the most conservative Democrats" \citep{wiki:Ben_Nelson}, it's not surprising that his voting record is closer to Republicans and has some connections with them. 
\begin{figure}
\centering
\includegraphics[scale=0.35]{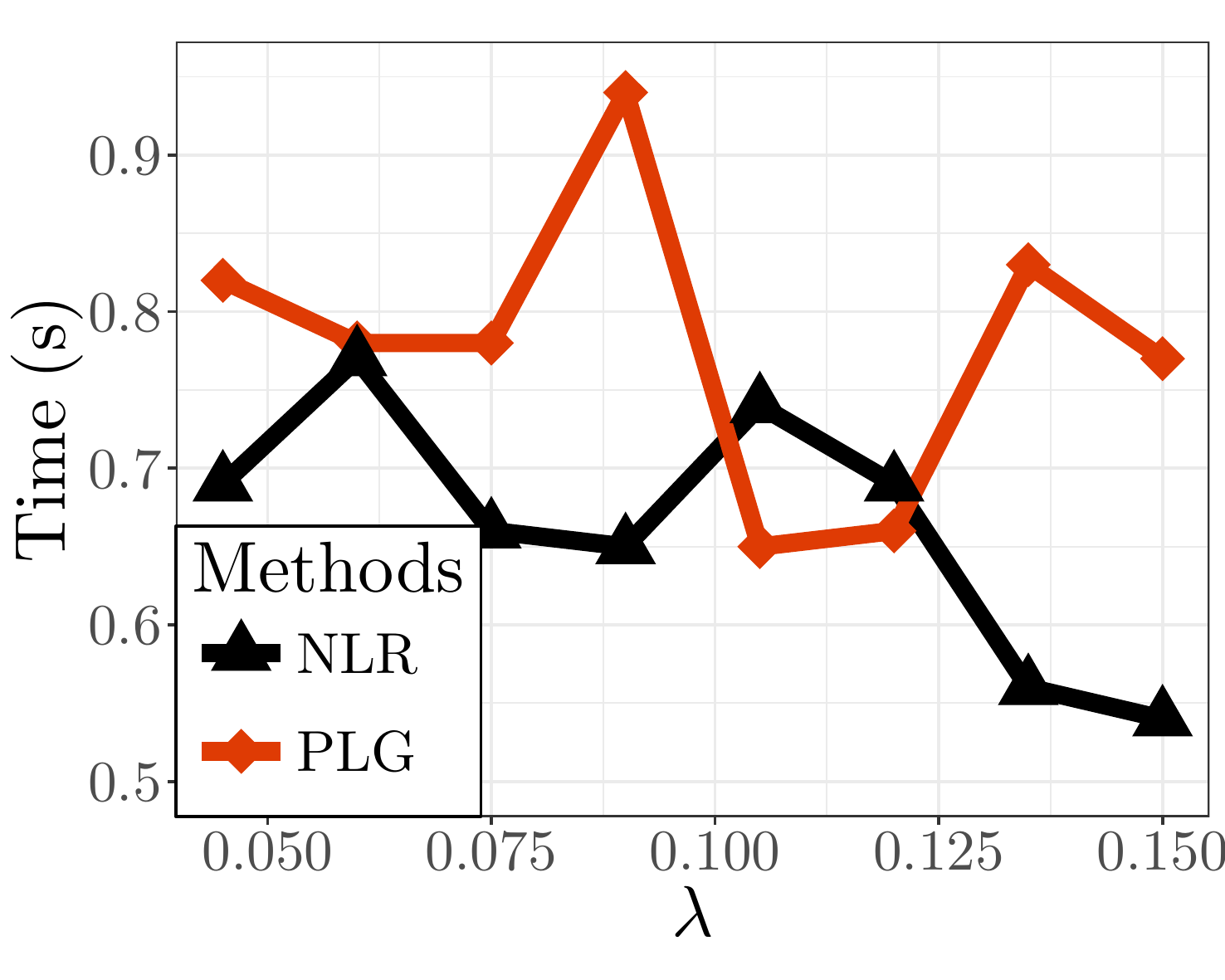}
\caption{Computation time for senator data using NLR and PLG under different $\lambda$ }
\label{fig:realTime} \vspace{-4mm}
\end{figure}
Furthermore, senators in the same parties and the same states tend to have more connections. These findings all coincide with conventional wisdom, suggesting that PLG can capture interesting dependencies in practical problems. 
We also contrast the efficiency of NLR and PLG on this real world data under different $\lambda$'s in Figure \ref{fig:realTime}. Again, the results of BMN are not included because of its long runtime. Similar to the results in Figure \ref{fig:time}, the computation time is almost the same for the two methods, indicating the high efficiency of PLG for practical problems.

\section{Conclusion}
For the task of accelerating the optimization of PL, by the equivalence between the objective functions of PL and LR we have studied an optimization method for PL models in the context of BPMNs. Experimental results suggest that PLG is a viable candidate towards scalable and efficient learning of BPMNs even in extreme conditions. Although we focus on binary pairwise Markov networks, our method is generally applicable to other discrete Markov networks whose pseudo-likelihood functions have a close relationship to logistic loss functions.  

\appendix

\bibliographystyle{named}
\bibliography{PLG}
\end{document}